\documentclass[conference]{IEEEtran}
\IEEEoverridecommandlockouts
\usepackage{cite}
\usepackage{amsmath,amssymb,amsfonts}
\usepackage{algorithmic}
\usepackage{graphicx}
\usepackage{textcomp}
\usepackage{xcolor}
\usepackage{booktabs}
\usepackage{multirow}
\usepackage{amsmath}
\usepackage{svg}
\usepackage{graphicx}
\usepackage{subcaption}
\usepackage{float}
\usepackage{comment}
\def\BibTeX{{\rm B\kern-.05em{\sc i\kern-.025em b}\kern-.08em
    T\kern-.1667em\lower.7ex\hbox{E}\kern-.125emX}}
\begin{document}

\title{A Novel Preprocessing-Driven Approach to Remaining Useful Life (RUL) Prediction Using Temporal Convolutional Networks (TCN)
}

\author{\IEEEauthorblockN{Florent Imbert}
\IEEEauthorblockA{\textit{Machine Learning Group,} \\
\textit{Luleå University of Technology,}\\
Luleå, Sweden \\
florent.imbert@associated.ltu.se}
\and
\IEEEauthorblockN{Tosin Adewumi}
\IEEEauthorblockA{\textit{Machine Learning Group,} \\
\textit{Luleå University of Technology,}\\
Luleå, Sweden \\
tosin.adewumi@ltu.se}
\and
\IEEEauthorblockN{Hui Han}
\IEEEauthorblockA{\textit{Machine Learning Group,} \\
\textit{Luleå University of Technology,}\\
Luleå, Sweden \\
hui.han@ltu.se}
}

\maketitle

\begin{abstract}
Accurate prediction of Remaining Useful Life (RUL) in aero-engines is vital for predictive maintenance, improved operational reliability, and reduced lifecycle costs. While deep learning approaches have demonstrated strong potential in this area, most existing methods focus primarily on model architecture design and treat input features uniformly, often neglecting the influence of data preprocessing.

In this work, we propose a novel preprocessing pipeline that enhances RUL prediction by improving data quality and temporal representation before model training. Our approach leverages complete temporal sequences and generates RUL estimates at each timestep, enabling the model to capture fine-grained degradation dynamics and deliver continuous prognostic insights throughout the engine’s operational life.

To validate the effectiveness of the proposed pipeline, we conduct experiments on the NASA C-MAPSS dataset. Comparative evaluations against a suite of state-of-the-art neural models including CNN, RNN, LSTM, DCNN, TCN, BiGRU-TSAM, AGCNN, and ATCN, demonstrate that our approach consistently achieves superior accuracy and robustness in aero-engine RUL prediction. These results highlight the critical role of preprocessing in maximizing the effectiveness of neural prognostic models.
\end{abstract}

\begin{IEEEkeywords}
Remaining useful life prediction, Temporal convolutional network, data preprocessing
\end{IEEEkeywords}

\section{Introduction}

Prognostics and Health Management (PHM) plays a pivotal role in modern industrial systems by enabling the assessment of system health and the prediction of impending failures. A central task in PHM is the estimation of Remaining Useful Life (RUL), defined as the time interval an asset is expected to operate before reaching a failure threshold. Accurate RUL prediction facilitates proactive maintenance scheduling, enhances system reliability, and reduces operational costs by minimizing unplanned downtime.

The development of data-driven RUL models has been greatly advanced by publicly available benchmark datasets, with the Commercial Modular Aero-Propulsion System Simulation (CMAPSS) dataset \cite{cmapss} from NASA being the most widely adopted. CMAPSS offers multivariate time-series sensor data from simulated aircraft engines subjected to varying operating conditions and degradation modes. It has become a de facto standard for evaluating prognostic algorithms in both academia and industry.

Recent research has predominantly concentrated on architectural innovations in deep learning such as Long Short-Term Memory (LSTM) networks, Gated Recurrent Units (GRU), Temporal Convolutional Networks (TCN), and Transformers to model temporal dependencies in RUL prediction. However, comparatively little attention has been paid to the upstream components of the learning pipeline, particularly the preprocessing and structuring of input data.

In this work, we shift the focus from model architecture to the data preparation process that precedes it. Using the Temporal Convolutional Network (TCN) as architecture, we investigate how variations in input formatting, feature scaling, temporal windowing, and target labeling affect model performance. Our approach highlights the critical impact of preprocessing in learning degradation patterns and improving prediction accuracy.

By emphasizing the role of data quality and representation, we aim to establish a set of best practices for constructing effective RUL prediction pipelines using the CMAPSS dataset. The proposed preprocessing-driven methodology not only enhances the performance of TCN but also offers insights that are transferable to other deep learning models in the PHM domain.

\section{Related Work}

Remaining Useful Life (RUL) prediction methods are commonly categorized into three main types: physical model-based, knowledge-based, and data-driven approaches \cite{vrignat2022sustainable}.

\subsection{Physical and Knowledge-Based Methods}
Physical model-based methods rely on simulating the underlying degradation mechanisms of components (e.g., crack propagation or corrosion dynamics) to estimate RUL \cite{ellis2022hybrid}. While highly interpretable, these approaches require domain-specific knowledge and often become infeasible for systems with complex or poorly understood failure modes \cite{sun2023lightweight}.

Knowledge-based methods use expert-defined rules or empirical degradation profiles to infer RUL from observed states \cite{wang2022research,djeziri2019data}. These methods can be effective when historical insights are abundant and well-structured. However, they suffer from limited generalizability and rely heavily on the quality of prior knowledge.

\subsection{Data-Driven Methods} 
With the proliferation of sensor data, data-driven methods have emerged as dominant solutions for RUL prediction. These methods can be divided into two subgroups: traditional statistical/machine learning models and deep learning models.

Statistical and classical machine learning approaches include regression models, random forests, and support vector machines \cite{wang2021multiscale,yu2022adaptive,namar2022start,soni2021multiclass}. These methods typically require extensive manual feature engineering to extract degradation indicators from high-dimensional time-series data. While interpretable, their performance is often limited by the quality of handcrafted features and sensitivity to distribution shifts in operating conditions \cite{cui2023two}.

Deep learning-based approaches aim to learn hierarchical feature representations directly from raw sensor inputs. Early work applied CNN to capture local temporal patterns in time-series data. For instance, \cite{babu2016deep} demonstrated a CNN-based framework for RUL estimation, exploiting convolutional layers for automatic feature extraction. However, CNN are constrained by limited receptive fields, which hinder their ability to model long-term dependencies.

To address this, \cite{li2018remaining} introduced a deeper DCNN with five convolutional layers, expanding the temporal receptive field and improving RUL prediction on the CMAPSS dataset, albeit at the cost of increased computational complexity.

Recurrent models, such as RNN \cite{zhang2019remaining} and LSTM \cite{wu2020data}, were introduced to better capture sequential dependencies. LSTM, with their gating mechanisms, alleviated the vanishing gradient issue and became widely adopted for RUL prediction. Yet, their sequential nature limits parallelization and training efficiency.

To improve both efficiency and sequence modeling capabilities, Temporal Convolutional Networks (TCN) have been proposed as an alternative. TCN combine dilated convolutions with residual connections to capture long-range temporal patterns in a parallelizable manner. \cite{ji2019remaining} applied TCN to RUL prediction, showing competitive performance and faster training compared to recurrent models.

Recent advances have focused on incorporating attention mechanisms to further improve feature relevance and temporal interpretability. For instance, \cite{yang2023bidirectional} combined BiGRU with a Temporal Self-Attention Mechanism (TSAM) to dynamically weigh time steps based on their contribution to RUL prediction. Similarly, \cite{liu2021remaining} introduced channel attention into a BiGRU framework to capture inter-sensor feature importance. Extending this idea, \cite{liu2022aircraft} applied channel attention prior to a Transformer model to enhance representation learning.

More recently, the Adaptive Temporal Convolutional Network (ATCN) \cite{zhang2024attention} integrated attention mechanisms into TCN, enabling the model to focus selectively on critical temporal segments during degradation progression.

Given its superior performance across multiple CMAPSS subsets, ATCN stands as a natural baseline for comparison in this study.
\cite{zhang2024attention}  approach combines two complementary attention mechanisms with a TCN backbone:
\begin{itemize}
    \item Improved Self-Attention (ISA): Positioned before the TCN, this mechanism enhances time steps that are more strongly correlated with RUL, allowing the model to focus on temporally relevant inputs, by replacing linear weight matrices (query matrix (Q), key matrix (K), and value matrix (V)) with 1×1 convolutions.
    \item Squeeze-and-Excitation (SE) Mechanism: Applied after the TCN, this module adaptively weights the importance of different sensor channels, emphasizing those most relevant to degradation trends.
\end{itemize}
On the preprocessing side, ATCN applies the following strategy:
\begin{itemize}
    \item Sensor selection: Only a subset of 14 informative sensors is retained.
    \item Min-max normalization: Each sensor channel is scaled to a [0, 1] range.
    \item Sliding window segmentation: Fixed-length windows (e.g., 31 time steps for FD001) are used to generate training samples, increasing data volume but introducing overlap.
    \item Piecewise linear RUL labeling: The RUL is capped at 125 cycles and decreases linearly beyond a certain point in time.
\end{itemize}
While ATCN achieves strong performance, especially under multi-operational conditions, our method explores an alternative path: instead of introducing additional attention layers, we improve RUL prediction by rethinking the preprocessing pipeline, using a TCN as model. This makes ATCN an ideal baseline to assess the impact of preprocessing choices independently of architectural complexity.

\section{Proposed methods}

Deep learning models such as LSTM, Transformers, and Temporal Convolutional Networks (TCN) have demonstrated strong performance in RUL prediction tasks by learning from sensor-based time-series data. However, while considerable attention has been devoted to advancing model architectures, comparatively little effort has been made to investigate how data preprocessing choices affect model performance and generalization.

Most existing works rely on fixed preprocessing routines such as standard windowing, normalization, and target labeling. These routines are often treated as a secondary concern despite their fundamental role in shaping the learning dynamics of the model, especially in domains where data is scarce or incomplete.

In this work, we propose a novel preprocessing pipeline for RUL prediction and a customized TCN model. The framework is designed to align the preprocessing pipeline with the temporal learning capabilities of TCN and address common limitations found in prior approaches. The proposed pipeline consists of the following key steps:
\begin{enumerate}
    \item \textbf{Training on complete sequences:} Instead of segmenting each engine trajectory into multiple overlapping sub-sequences using sliding windows, the model is trained directly on full-length sequences.
This provides a comprehensive view of the entire degradation process, allowing the model to learn global behavioral patterns and full lifecycle dynamics.
    \item \textbf{Standardized normalization across features:} Sensor features are standardized using z-score normalization to ensure zero mean and unit variance. This helps mitigate the influence of differing scales across features and avoids introducing distributional biases during training.
    \item \textbf{Clipping the maximum RUL to 125 cycles:} RUL values are capped at a maximum of 125 cycles, in line with common practice in the literature. This reduces label variance in early lifecycle stages and facilitates comparison with existing methods.
    \item \textbf{Random trimming of sequence ends:} Each sequence is truncated by a random number of time steps near the end of its lifecycle, while retaining at least 30 observations. This simulates realistic use cases where failure is not always observed and enhances the model's ability to generalize to partial lifecycles.
    \item \textbf{Prediction of RUL at each time point:} The model is trained to produce a RUL estimate at every time step of the input sequence. This allows the model to learn localized degradation patterns by leveraging temporal context throughout the entire sequence.
\end{enumerate}

By incorporating these preprocessing steps, the proposed framework aims to improve the robustness and generalizability of RUL prediction models. Using the CMAPSS dataset, we evaluate the impact of this new pipeline and demonstrate how thoughtful preprocessing can lead to more accurate and stable predictions across different operating conditions and fault modes.

\subsection{CMAPSS dataset}
This study utilizes the CMAPSS (Commercial Modular Aero-Propulsion System Simulation) dataset, developed by NASA. It is widely recognized as a benchmark for evaluating RUL prediction models in aviation maintenance and condition monitoring. The dataset provides multivariate time-series data collected from simulated aircraft engines, including sensor measurements and operational settings.

CMAPSS is divided into four subsets each comprising a training and test set. These subsets differ in terms of operational conditions and fault modes. The number of data and specific operational conditions are detailed in~\ref{CMAPSS}.

\begin{table}[h!] 
\footnotesize 
\centering
\caption{CMAPSS dataset and subsets. Table taken from \cite{zhang2024attention}}
\label{CMAPSS}
\begin{tabular}{|l|c|c|c|c|}
\hline
\textbf{Subsetet} & \textbf{FD001} & \textbf{FD002} & \textbf{FD003} & \textbf{FD004} \\
\hline
Train samples & 100 & 260 & 100 & 249 \\
\hline
Test samples  & 100 & 259 & 100 & 248 \\
\hline
Operation conditions  & 1   & 6   & 1   & 6   \\
\hline
Fault modes           & 1   & 1   & 2   & 2   \\
\hline
\end{tabular}
\end{table}
The training sets for FD001, FD002, FD003, and FD004 contain 100, 260, 100, and 249 engine trajectories, respectively. Each trajectory represents the full operational lifespan of a single engine until failure. These trajectories include time-series data of 21 sensor measurements, three operational condition parameters, and associated serial numbers.

The corresponding testing sets include 100, 259, 100, and 248 trajectories for FD001 through FD004, respectively. Unlike the training data, the testing trajectories capture engine behavior only up to a specific point prior to failure, without reaching the full degradation cycle. The details of the 21 monitored engine sensors are presented in Table~\ref{CMAPSS_D}.

\begin{table}[h!] 
\footnotesize 
\centering
\caption{Engine sensors description. Table taken from \cite{zhang2024attention}}
\label{CMAPSS_D}
\begin{tabular}{|c|l|l|l|}
\hline
\textbf{Number} & \textbf{Symbols} & \textbf{Description} & \textbf{Units} \\
\hline
1  & T2         & Total temperature at fan inlet            & $^\circ$R    \\
\hline
2  & T24        & Total temperature at LPC outlet           & $^\circ$R    \\
\hline
3  & T30        & Total temperature at HPC outlet           & $^\circ$R    \\
\hline
4  & T50        & Total temperature at LPT outlet           & $^\circ$R    \\
\hline
5  & P2         & Pressure at fan inlet                     & psia         \\
\hline
6  & P15        & Total pressure in bypass-duct             & psia         \\
\hline
7  & P30        & Total pressure at HPC outlet              & psia         \\
\hline
8  & Nf         & Physical fan speed                        & rpm          \\
\hline
9  & Nc         & Physical core speed                       & rpm          \\
\hline
10 & Epr        & Engine pressure ratio (P50/P2)            & -            \\
\hline
11 & Ps30       & Static pressure at HPC outlet             & psia         \\
\hline
12 & phi        & Ratio of fuel flow to Ps30                & pps/psi      \\
\hline
13 & NRf        & Corrected fan speed                       & rpm          \\
\hline
14 & NRc        & Corrected core speed                      & rpm          \\
\hline
15 & BPR        & Bypass ratio                              & -            \\
\hline
16 & farB       & Burner fuel–air ratio                     & -            \\
\hline
17 & htBleed    & Bleed enthalpy                            & -            \\
\hline
18 & Nf\_dmd     & Demanded fan speed                        & rpm          \\
\hline
19 & PCNfR\_dmd  & Demanded corrected fan speed              & rpm          \\
\hline
20 & W31        & HPT coolant bleed                         & lbm/s        \\
\hline
21 & W32        & LPT coolant bleed                         & lbm/s        \\
\hline
\end{tabular}
\end{table}

\begin{figure}[h!]
    \centering
    \includegraphics[width=0.73\columnwidth]{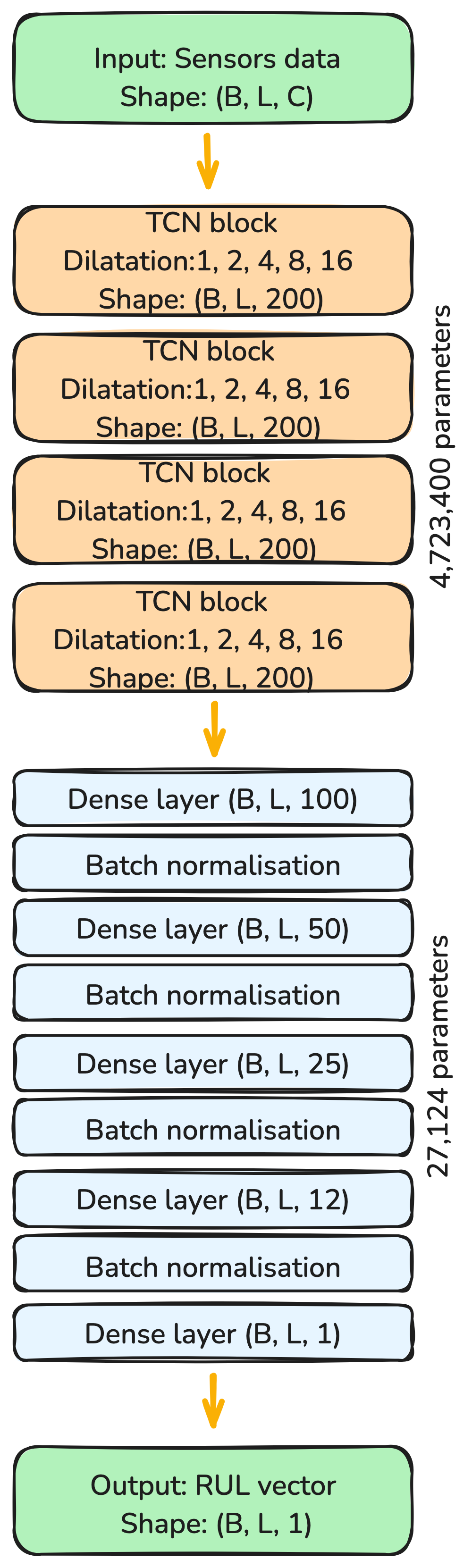}
    \caption{Proposed TCN model, with a receptive field 125. Where: B = batch size, L = sequence length, C = channels}
    \label{fig:TCN}
\end{figure}

\subsection{A TCN-based approach}

We adopt a TCN-based architecture for RUL prediction, selected for its robustness, training efficiency, and superior temporal modeling capabilities. Compared to recurrent networks such as LSTM, TCN avoid vanishing gradient issues and benefit from parallelism. Unlike Transformer models, TCN are computationally lighter and require less data to generalize effectively, an advantage in settings with limited data. Moreover, recent observations have highlighted the superior performance of this architecture in processing time series, especially in sensor-based trajectory reconstruction \cite{swaileh2023online, IMBERT2025111231}.

\begin{figure}[h!]
    \centering
    \includegraphics[width=0.9\columnwidth]{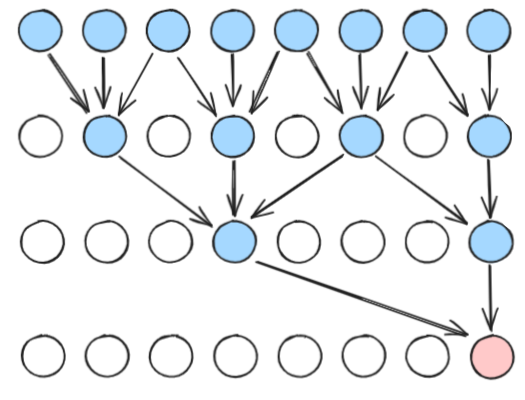}
    \caption{TCN concept dilated convolutions to extract temporal features}
    \label{fig:rf}
\end{figure}

The architecture that we have designed (Fig.\ref{fig:TCN}) consists of four stacked residual TCN blocks (Fig.\ref{fig:rf}), each comprising five dilated 1D convolutions with dilation rates [1, 2, 4, 8, 16], a kernel size of 3, a number of filter of 200 and dropout rate of 0.3. Each convolutional layer is followed by batch normalization and ReLU activation. The final output is processed through a five-layer projection head with progressively reduced dimensionality, producing a single RUL estimate per time step.

\subsection{A new preprocessing chain for RUL}
\begin{figure*}[h!]
    \centering
    \includegraphics[width=0.9\textwidth]{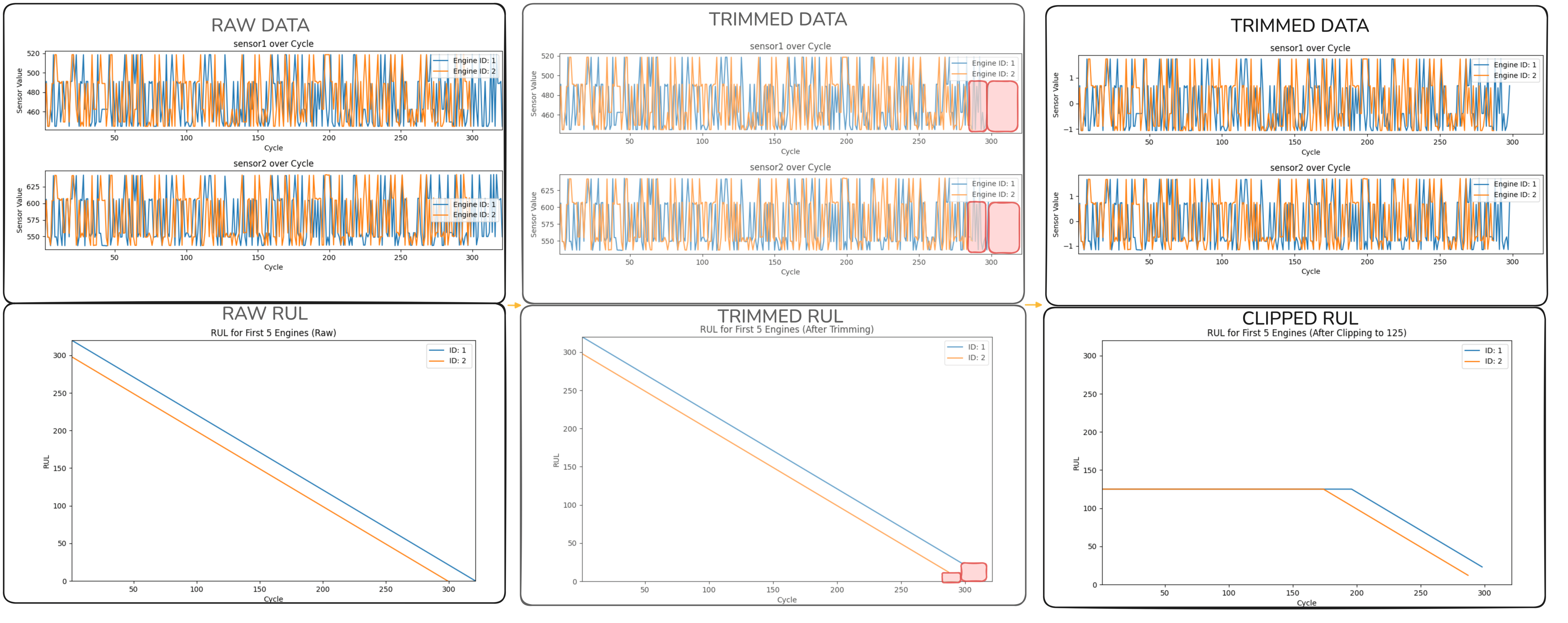}
    \caption{Preprocessing pipeline for RUL. First column raw data, second column trimed times series, third column clip RUL}
    \label{fig:preprocessing}
\end{figure*}

To align preprocessing with the temporal modeling strengths of the TCN, we introduce a novel data pipeline (Fig.~\ref{fig:preprocessing}) that incorporates five core components: training on full sequences, random end trimming, standardized normalization, RUL clipping, and dense per-timestep labeling.

\subsubsection{Avoiding Sliding Window Bias}
\begin{figure}[H]
    \centering
    \includegraphics[width=1.0\columnwidth]{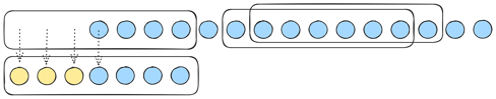}
    \caption{TCN sliding window with zero padding on edges}
    \label{fig:windows}
\end{figure}

Many prior methods, such as \cite{zhang2024attention}, use sliding windows to generate fixed-length samples, greatly inflating the number of training examples. However, this also increases padding at the boundaries of each window. This introduces repetitive noise and reduces the effectiveness of the temporal convolution operations (Fig.~\ref{fig:windows}) of the TCN. The TCN operates like a sliding window. Expanding the time series also amplifies the padding at the edges of the predictions.

In contrast, we propose training directly on the full-length engine trajectories without segmentation. This preserves the integrity of long-range temporal dependencies and reduces bias introduced by excessive padding. For instance, in the FD001 subset, this approach maintains the original 100 trajectories rather than expanding to over 17,000 samples (For the FD001 subdataset of the CMAPSS dataset).

\subsubsection{Random End Trimming for Generalization}
To simulate real-world scenarios where failures are not always fully observed, we randomly remove between 10 and 75 time steps from the end of each sequence, ensuring a minimum length of 30. This encourages the model to learn degradation patterns throughout the sequence, rather than overfitting to final-stage cues (i.e., when RUL = 0), thereby improving generalization in early warning tasks.

\subsubsection{Standardization Across Features}
Input features are standardized to zero mean and unit variance using z-score normalization:

\begin{equation}
x' = \frac{x - \mu}{\sigma} \\
\label{stand}
\end{equation}

\begin{equation*}
\begin{aligned}
\text{Where} \\
x &\text{ is the original value,} \\
\mu &\text{ is the mean of the feature,} \\
\sigma &\text{ is the standard deviation of the feature,} \\
x' &\text{ is the standardized value.}
\end{aligned}
\end{equation*}

This prevents features with larger numerical ranges from dominating the learning process and avoids sensitivity to outliers, which is a limitation of min-max scaling. Standardization enables the model to more effectively learn degradation trends across varying sensors and conditions.

\subsubsection{RUL Clipping Strategy}
Consistent with prior work \cite{li2018remaining, liu2021remaining, liu2022aircraft, zhang2024attention}, we cap RUL values at a maximum of 125 cycles. This reduces label variance during the initial stable phase of operation and improves convergence stability during training. The clipped piecewise RUL function is defined as:

\begin{equation}
y_t = 
\begin{cases}
R_{\text{max}}, & t \leq t_0 \\
R_{\text{max}} - k(t - t_0), & t > t_0
\end{cases}
\end{equation}

\begin{equation*}
\begin{aligned}
\text{Where} \\
t &\text{ represents the time step,} \\
t_0 &\text{ is the time at which degradation starts,} \\
R_{\text{max}} &\text{ is the maximum RUL (set to 125 cycles),} \\
k &\text{ is a constant that determines the rate} \\
&\text{of degradation once the failure starts.}
\end{aligned}
\end{equation*}


\subsubsection{Prediction of RUL at each time point}
Rather than predicting RUL only at selected time points, our model is trained to predict RUL at every time step. This dense supervision enables the TCN to exploit both local and global temporal contexts, enhancing its ability to detect degradation onset and understand transient behaviors in the sensor data.

\section{Experimental setting}

\subsection{Evaluation metrics}
To evaluate the performance of the proposed method, we employ two key evaluation metrics: Root Mean Square Error (RMSE) and the Scoring Function (Score), the latter introduced by \cite{saxena2008damage}. RMSE quantifies the average deviation between the predicted and actual values and is defined in Equation~\eqref{eq:rmse}. 

\begin{equation}
\text{RMSE} = \sqrt{ \frac{1}{N} \sum_{n=1}^{N} (y_n - \hat{y}_n)^2 }
\label{eq:rmse}
\end{equation}

\begin{equation*}
\begin{aligned}
\text{Where} \\
N & \text{ denotes the total number of samples,} \\
y_n &\text{ represent the true RUL of the \(n\)-th instance,} \\
\hat{y}_n &\text{ represent the predicted RUL of the \(n\)-th instance.}
\end{aligned}
\end{equation*}


The Score, defined in Equation~\eqref{eq:score}, introduces asymmetric penalties on prediction errors, with a heavier penalty imposed on late predictions.

\begin{equation}
\text{Score} =
\begin{cases}
\sum_{n=1}^{N} \left( e^{-\frac{d_n}{13}} \right), & \text{if } d_n < 0 \\
\sum_{n=1}^{N} \left( e^{\frac{d_n}{10}} \right), & \text{if } d_n \geq 0
\end{cases}
\label{eq:score}
\end{equation}

\begin{equation*}
\begin{aligned}
\text{Where} \\
d_n = \hat{y}_n - y_n  & \text{ is the error for the \(n\)-th sample.} 
\end{aligned}
\end{equation*}


The main difference between RMSE and Score lies in how they handle prediction bias. RMSE treats early and late predictions equally, penalizing all deviations symmetrically. In contrast, the Score metric assigns a greater penalty to late predictions, which aligns with the practical requirement for early warnings in predictive maintenance applications. Early predictions enable timely intervention and maintenance scheduling, whereas late predictions could result in unexpected system failures. However, the Score is more sensitive to outliers, where large prediction errors may significantly impact the overall score. Therefore, RMSE and Score are commonly used together to provide a more comprehensive evaluation of the model's predictive performance.

\subsection{Related works comparison}
The proposed method, which incorporates a new preprocessing pipeline and a Temporal Convolutional Network (TCN) model, is validated by comparing its experimental results with those of various other deep learning approaches. 

These methods include Convolutional Neural Networks (CNN) \cite{babu2016deep}, Recurrent Neural Networks (RNN) \cite{li2018remaining}, Long Short-Term Memory networks (LSTM) \cite{li2018remaining}, Deep Convolutional Neural Networks (DCNN) \cite{li2018remaining}, Temporal Convolutional Networks (TCN) \cite{ji2019remaining}, Bidirectional Gated Recurrent Units with Temporal Self-Attention Mechanism (BiGRU-TSAM) \cite{zhang2022prediction}, Channel-Attention based BiGRU and CNN (AGCNN) \cite{liu2021remaining}, Channel Attention combined with Transformer \cite{liu2021remaining}, and the Attention-enhanced TCN (ATCN) model \cite{zhang2024attention}. 

All experiments are performed under consistent experimental settings to ensure a fair comparison: batch size is 8, learning rate is 0.01, number of epochs is 1000, with an early stop patience of 40. The performance of each method is evaluated using two key metrics: Root Mean Squared Error (RMSE) and Score, both of which provide insights into the accuracy and effectiveness of the respective models.

\section{Results}

\subsection{Quantitative Evaluation}
We present quantitative results in Table~\ref{tab:rmse_score}, comparing our proposed method combining a TCN model with a novel preprocessing pipeline against several state-of-the-art approaches. To ensure robustness and account for model variability, we perform 10 independent training runs for each FD dataset and report the mean and standard deviation (SD) of both RMSE and Score metrics.

Our method achieves the lowest average RMSE across all datasets and outperforms all baselines in Score on FD002 and FD004. Although the Score metric on FD001 and FD003 is slightly higher than that of ATCN~\cite{zhang2024attention}, our performance remains competitive.

A key factor contributing to this improvement is the proposed preprocessing strategy, which enhances data clarity and structure. By training on full sequences and incorporating realistic trimming and label constraints, the model gains a stronger understanding of lifecycle dynamics. This leads to better generalization and more stable predictions across diverse operating conditions.

It is worth noting that prediction errors for FD002 and FD004 are significantly higher compared to FD001 and FD003 across all models. This disparity can be attributed to the fact that FD002 and FD004 involve multiple operating conditions and fault modes, whereas FD001 and FD003 are restricted to a single operational setting. This added complexity presents a greater challenge for generalization, making our model’s performance on these datasets particularly notable.

\begin{table*}[!h]
\centering
\caption{The comparative results (RMSE and score) of the proposed method against previous ones.}
\label{tab:rmse_score}
\begin{tabular}{|l|c|c|c|c|c|c|c|c|}
\hline
\multirow{2}{*}{\textbf{Methods}} & \multicolumn{2}{c|}{\textbf{FD001}} & \multicolumn{2}{c|}{\textbf{FD002}} & \multicolumn{2}{c|}{\textbf{FD003}} & \multicolumn{2}{c|}{\textbf{FD004}} \\
\cline{2-9}
 & \textbf{RMSE} & \textbf{Score} & \textbf{RMSE} & \textbf{Score} & \textbf{RMSE} & \textbf{Score} & \textbf{RMSE} & \textbf{Score} \\
\hline
RNN                        & 13.44 & 339.2   & 20.43 & 14245   & 13.36 & 315.7   & 24.02 & 13931   \\
\hline
LSTM                       & 13.52 & 431.7   & 22.64 & 14459   & 13.54 & 347.3   & 24.21 & 14322   \\
\hline
BiGRU-TSAM                 & 12.56 & 213.35  & 19.04 & 2264.13 & 12.45 & 232.86  & 20.34 & 3610.34 \\
\hline
CNN                        & 18.45 & 1286.7  & 30.29 & 13570   & 19.82 & 1596.2  & 29.16 & 7886.4  \\
\hline
AGCNN                      & 12.40 & 225.51  & 19.01 & 1492.76 & 13.39 & \textbf{227.09} & 19.84 & 3392.6  \\
\hline
DCNN                       & 12.54 & 237.7   & 24.24 & 12824   & 12.43 & 284.4   & 21.94 & 14266   \\
\hline
Channel attention + Transformer & 12.25 & 198   & 20.8 & 1575    & 12.39 & 209     & 19.86 & 1741    \\
\hline
TCN                        & 13.51 & 266     & 22.43 & 2831    & 14.41 & 401     & 21.93 & 2321    \\
\hline
ATCN             & 11.48 & \textbf{194.25} & 18.52 & 1210.57 & 11.34 & 249.19 & 17.8 & 1934.86 \\
\hline
\textbf{Ours (mean)} 
& \textbf{9.26} & 376.67 
& \textbf{11.04} & \textbf{834.50} 
& \textbf{7.16} & 362.10 
& \textbf{11.42} & \textbf{1717.09} \\
\hline
\textbf{Ours ( ± SD)} 
& \textbf{± 0.96} & ± 116.79 
& \textbf{± 0.81} & \textbf{± 230.46} 
& \textbf{± 0.46} & ± 63.62 
& \textbf{± 0.72} & \textbf{± 319.88} \\
\hline
\end{tabular}
\end{table*}

\subsection{Qualitative Evaluation}
\begin{figure*}[ht]
    \centering
    \begin{subfigure}[b]{0.49\textwidth}
        \centering
        \includegraphics[width=\linewidth]{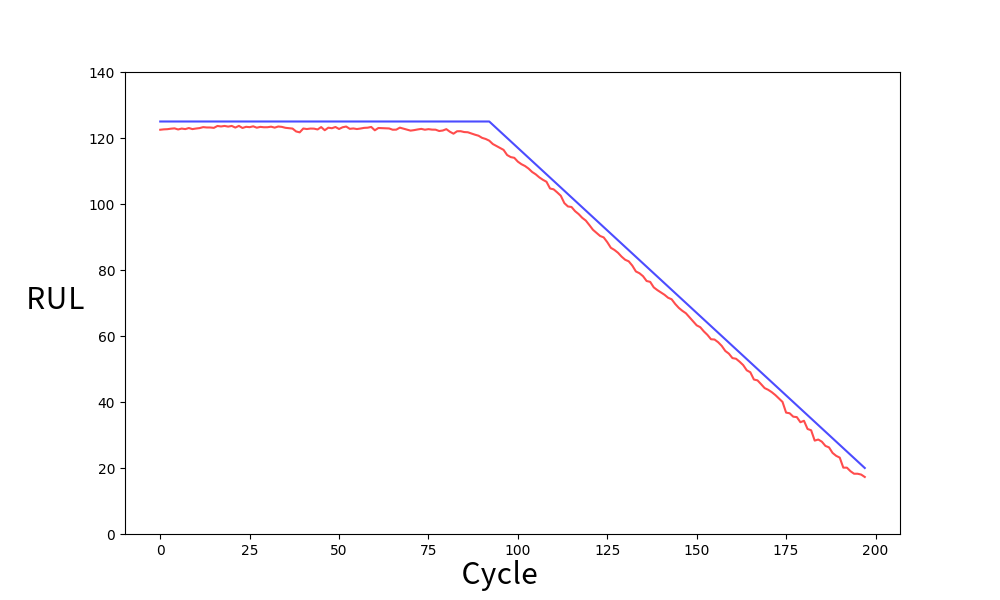}
        \caption{Test engine unit 100 in FD001}
        \label{fig:FD001}
    \end{subfigure}
    \hfill
    \begin{subfigure}[b]{0.49\textwidth}
        \centering
        \includegraphics[width=\linewidth]{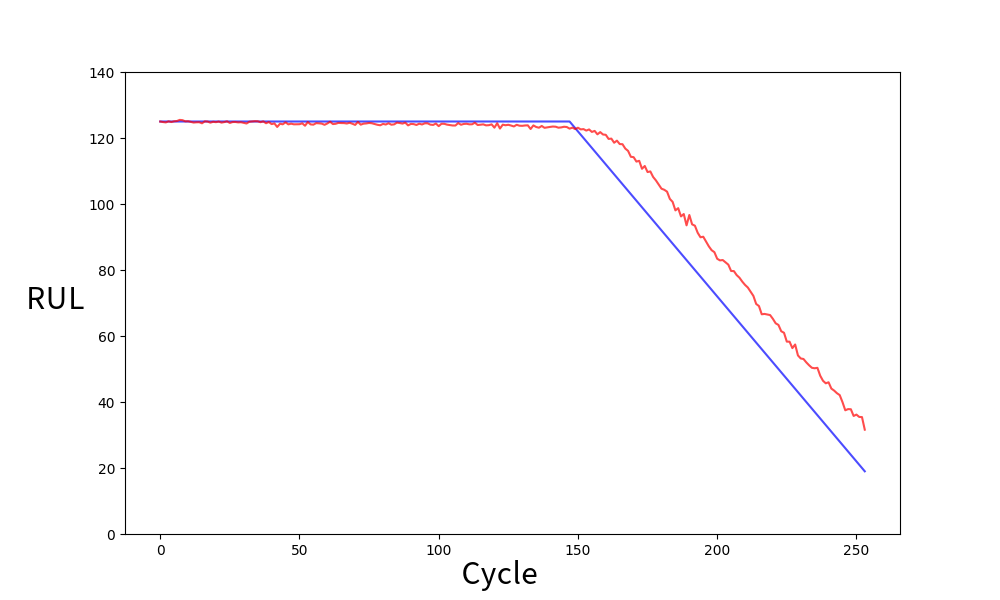}
        \caption{Test engine unit 185 in FD002}
        \label{fig:FD002}
    \end{subfigure}

    \vskip\baselineskip

    \begin{subfigure}[b]{0.49\textwidth}
        \centering
        \includegraphics[width=\linewidth]{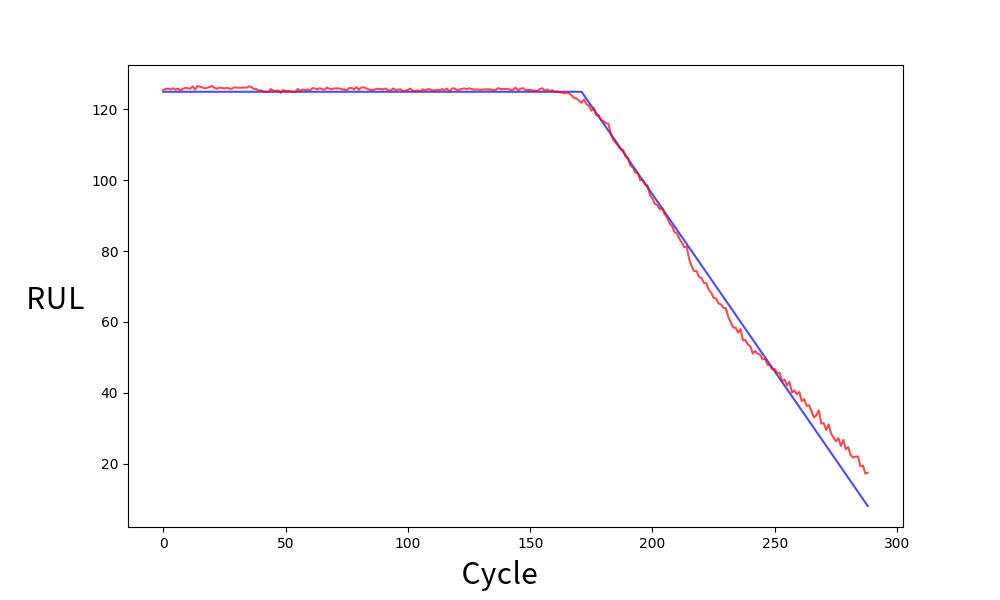}
        \caption{Test engine unit 99 in FD003}
        \label{fig:FD003}
    \end{subfigure}
    \hfill
    \begin{subfigure}[b]{0.49\textwidth}
        \centering
        \includegraphics[width=\linewidth]{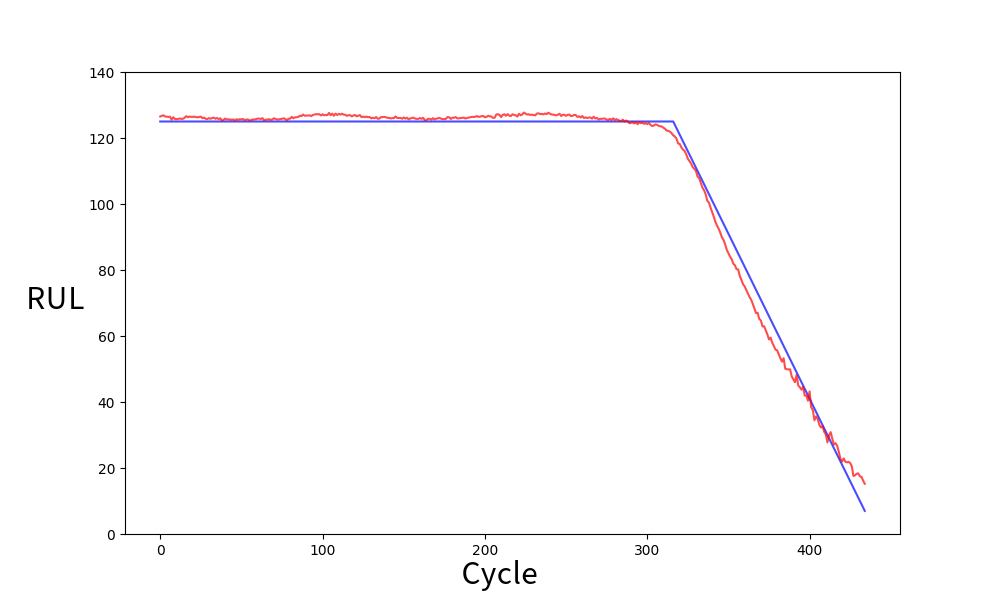}
        \caption{Test engine unit 135 in FD004}
        \label{fig:FD004}
    \end{subfigure}
    
    \caption{RUL/cycle prediction (red) vs actual RUL (blue) results on test engine units for FD001 to FD004.}
    \label{fig:result}
\end{figure*}

For qualitative assessment, we reuse the visual comparison format from \cite{zhang2024attention}, focusing on test engine units with diverse RUL profiles across FD001 to FD004 (Figure~\ref{fig:result}). Figure ~\ref{fig:result} illustrates the comparison between predicted Remaining Useful Life (RUL) and actual RUL for selected test engine units across the four CMAPSS subsets. In each subfigure, the predicted RUL (in red) is plotted against the actual RUL (in blue) over the engine's operational cycles. The results demonstrate that the prediction model accurately captures the degradation patterns of the engines, with strong alignment observed across all datasets.

Compared to prior methods, our approach produces RUL curves with noticeably fewer fluctuations and improved alignment with actual values.

This stability stems from training the model on complete RUL sequences. Exposure to full trajectories allows the TCN to learn the general shape of RUL curves, which typically exhibit an initial plateau followed by a linear degradation. This contrasts with models trained on windowed sub-sequences, which often struggle to capture long-range dependencies and global lifecycle structure.

Despite overall improvements, we observe two common types of prediction errors:
\begin{enumerate}
    \item Incorrect estimation of the RUL clipping value, where predictions prematurely flatten at the capped RUL (125 cycles).
    \item Incorrect estimation of the degradation onset, where the model fails to precisely predict the point at which RUL begins to decrease.
\end{enumerate}

These failure modes suggest opportunities for further enhancement, such as adaptive RUL capping or dynamic detection of degradation onset via auxiliary supervision.

\section{Conclusion}

In this study, we introduced a novel preprocessing pipeline coupled to a TCN model for Remaining Useful Life (RUL) prediction of aero-engines, shifting the focus from architectural innovations to optimizing the data preprocessing pipeline. Our approach emphasizes training on full-length sequences, applying random trimming to enhance generalization, standardizing sensor features for consistency, and adopting a capped RUL labeling strategy to reduce variance and improve learning stability.

By leveraging these tailored preprocessing strategies, we enabled Temporal Convolutional Networks (TCN) to more effectively capture degradation trends and lifecycle dynamics. Our method was evaluated on the CMAPSS dataset across all four sub-datasets (FD001–FD004). Results averaged over 10 independent runs show that our method achieves strong and consistent performance: RMSE of 9.26, 11.04, 7.16, and 11.42, and Score values of 376.67, 834.50, 362.10, and 1717.09, respectively. These results surpass prior state-of-the-art models on most CMAPSS sub-datasets while demonstrating low variability across runs, highlighting the robustness of the proposed preprocessing pipeline.

This work underscores the critical role of data representation in prognostic modeling. It demonstrates that substantial performance gains can be achieved without altering model architectures simply by aligning data preparation more closely with the learning capabilities of the model. Future work will explore dynamic data augmentation strategies, domain adaptation for unseen operating conditions, and integrating uncertainty quantification to enhance reliability and applicability in real-world predictive maintenance scenarios.

\section*{Acknowledgment}
The research of Hui Han is supported by the Wallenberg AI, Autonomous Systems and Software Program (WASP) funded by the Knut and Alice Wallenberg Foundation. Florent Imbert is supported by the Kempestiftelserna grant CSMK23-0109. The computations were enabled by the Berzelius resource at the National Supercomputer Center.

\bibliographystyle{IEEEtran}
\bibliography{bib}

\end{document}